\documentclass[runningheads]{llncs}
\usepackage{graphicx}
\usepackage{amsmath}
\usepackage{amssymb}
\usepackage{svg}
\usepackage{pdfpages}
\begin{document}
\title{Spot the Bot: Distinguishing Human-Written and Bot-Generated Texts Using Clustering and Information Theory Techniques}
\titlerunning{Spot the Bot: Distinguishing Human-Written and Bot-Generated Texts}
\author{Vasilii Gromov\inst{1}\orcidID{0000-0001-5891-6597}, Quynh Nhu {Dang}\inst{1}\orcidID{0000-0003-0450-7063}}
\authorrunning{V. Gromov \and Q.N. Dang}
\institute{National Research University Higher School of Economics, Moscow, Russia
\email{stroller@rambler.ru, dqnhu00@gmail.com}}
\maketitle              % typeset the header of the contribution
\begin{abstract}
With the development of generative models like GPT-3, it is increasingly more challenging to differentiate generated texts from human-written ones. There is a large number of studies that have demonstrated good results in bot identification. However, the majority of such works depend on supervised learning methods that require labelled data and/or prior knowledge about the bot-model architecture. In this work, we propose a bot identification algorithm that is based on unsupervised learning techniques and does not depend on a large amount of labelled data. By combining findings in semantic analysis by clustering (crisp and fuzzy) and information techniques, we construct a robust model that detects a generated text for different types of bot. We find that the generated texts tend to be more chaotic while literary works are more complex. We also demonstrate that the clustering of human texts results in fuzzier clusters in comparison to the more compact and well-separated clusters of bot-generated texts.

\end{abstract}
\section{Introduction}
{
    With the development of NLP methods it has become increasingly more difficult to distinguish computer-generated texts from human literature. Many advances have been made in bot detection in various fields. However, state-of-the-art solutions are obtained using supervised methods and depend heavily on labelled data. Not as many works concentrate on self-supervised or unsupervised learning and those that do usually deal with particular bots. Our main objective is to conduct a careful study of semantic paths of both literature and bot-generated texts to find a black-box method for spotting bots. The goal is to find a procedure that distinguishes human-written texts from bot-generated texts without prior knowledge about the bot.

    Our study provides a general view on how human-written texts and bot-generated texts differ on a semantic level and studies the compactness, separability and noisiness of clusters, as well as the types of text series (deterministic/chaotic/stochastic). Our hypothesis is that these characteristics should differ for human-written and bot-generated texts, and the findings can be used to create an algorithm for bot identification. The advantage of this algorithm lies in its universality and its ability to work with bots of different types — from simple Recurrent Neural Network models to more advanced GPT bots. Our study has shown that different methods highlight various properties of the semantic space. The analysis of the characteristics of semantic paths has shown that human-written texts are more complex, while the bot-generated texts tend to be simpler and more chaotic. The clustering of data has resulted in more compact and well-separated clusters for bot-generated texts and fuzzier clusters for human-written texts. The rest of this paper is organized as follows. In the next section we review recent advances in the bot detection field. Section 3 outlines the methods we have used for the analysis of semantic space. Section 4 provides the description of conducted experiments and presents the results. In Section 5 we give our conclusions.
}

\section{Literature Review}
{
    Recent years have seen a surge of interest in the bot detection task. Most studies employ feature-based supervised learning algorithms and centre around constructing features which are then used to build a classification model. There are a variety of methods to build such features. \cite{kang2012chatting} use simple lexical and syntactic features like letter frequency or average word length. \cite{heidari2021empirical} derive sentiment qualities of English and Dutch tweets by calculating their polarity. \cite{cardaioli2021sa} model a Twitter user through a set of stylistic features and distinguish bots from human accounts by analysing the consistency of their post style. \cite{chakraborty2022detection} combine text feature engineering and graph analytics. Similarly, \cite{dickerson2014using} propose SentiBot, an architecture that combines graph-based and sentiment and semantic analysis techniques. In our study, we focus on unsupervised machine learning algorithms, rather than supervised learning methods, and engineer features by clustering texts, examining the resulting semantic space and extracting various characteristics.

    Other approaches are based on information theory. \cite{chu2012detecting} characterise the differences between bot and human activity on Twitter by calculating the entropy of account activity statistics. They have found that humans have higher entropy than bots, which highlights their more complex timing behaviour. In our work we apply similar ideas to semantic trajectories of text data instead of meta-data. In \cite{gromov2017language} the authors study a natural language as an integral whole and ascertain that it is a self-organised critical system, whereas a separate literature text is ‘an avalanche’ in a semantic space. The latter fact further reinforces the argument for considering a trajectory in a semantic space as a unified object.
}
\section{Methodology}
{
    \subsection{Data}
    {
    \begin{table}
    \centering
    \caption{\label{corpora-details}
    Literature corpora details.
    }
    \begin{tabular}{lccc}
    \hline
    \textbf{language} & \textbf{corpus size} & \textbf{unique bigrams} & \textbf{library}\\
    \hline
    English & 11008 & 8m & spacy\\
    Russian & 12692 & 3m & natasha\\
    Vietnamese & 1071 & 6m & pyvi\\
    \hline
    \end{tabular}
    \end{table}
    For the human written corpus, the literary books were obtained via open sources. See Table~\ref{corpora-details} for corpora details and python libraries used for each language.
    To obtain bot-generated texts two models were utilised — a simple Long Short Term Memory recurrent neural network (LSTM), and a GPT-3. We use different models in order to design a working identification algorithm on both simple and complex bots. We train the LSTM model on subsets from the literary corpora and select pretrained GPT models from the huggingface database. To generate texts, for every 500th word from a literary piece we generate a text abstract of 500 words (the conventional size of a book page), therefore, the texts are generated of similar lengths as literary texts.
    }
    \subsection{Embeddings}
    {
    Word embeddings are obtained using the SVD of a document-term matrix \cite{bellegarda2007latent} and the word2vec models \cite{mikolov2013efficient}. The decision to use these two techniques was based on their semantic properties – both SVD and word2vec embeddings capture the structural relationships between words. In order to study word order correlations, we split the texts into n-grams and obtain final embeddings by concatenating word embeddings for each word in an n-gram. The collection of n-gram embeddings for each text is further referred to as a semantic path.
    }
    \subsection{Clustering}
    To analyse the semantic space, we use Wishart (density-based) \cite{wishart1969numerical} and K-Means \cite{macqueen1967classification} clustering techniques\footnote{Each algorithm has its advantages --- K-Means separates spherical clusters well, whereas Wishart algorithm does not make any assumptions about cluster shapes.}. We additionally explore fuzzy implementations of these algorithms to allow for the noisiness and imprecise nature of real-life data. We consider two algorithms: fuzzy clustering C-Means, \cite{bezdek1984fcm}, which is similar to K-Means, and Wishart clustering on fuzzy numbers.

    To fuzzify the data, we use the notion of fuzzy numbers with trapezoidal membership functions\cite{novak2012mathematical}. For each $j$-th component of an $m$-dimensional object $x$ we define the value for the fuzzy membership function as $\mu_j (x_j) = \frac{n_j}{\max\limits_j n_j}$, where $n_j$ is the normalised frequency of j-th component in the text. With fixed parameter values of $l_j, r_j, \Delta c = m_{2j}-m_{1j}$ we construct the fuzzy number. The ordered set of fuzzy numbers for each component of $x$ is the fuzzification of $x$.
    
    To fuzzify n-grams, we join fuzzifications of the words from n-grams accordingly to the fuzzy logic, i.e. take the minimum of fuzzy number membership functions. Finally, to use Wishart clustering algorithm (which only requires pairwise distances) on fuzzy data, we calculate the fuzzy distance as defined in \cite{novak2012mathematical}.
    
    \subsection{Entropy-complexity plane}
    The second method proposed in \cite{rosso2007distinguishing} distinguishes chaotic semantic paths from deterministic and stochastic ones. In order to test our hypothesis that the bot-generated texts are less complex and more chaotic, we calculate complexity and entropy measures of the word permutations. The position of the point in relation to the lower and upper theoretical boundaries points to the type of the series in question. Namely, simple deterministic processes occupy the bottom left corner of the plane, stochastic processes, the bottom right corner, whereas chaotic (complex deterministic) processes occupy areas adjacent to the vertex of the upper curve \cite{rosso2007distinguishing}. We also propose a modified variation for multidimensional use: for $m$-dimensional time series $(x^{t})_{t=1}^{L}$, $x_t \in \mathbb{R}^{m}$ for each of $m$ components of an n-gram we obtain permutation $\pi_d$ as in one-dimensional case. For multidimensional case we define the final permutation as $\Pi = (\pi_1,\pi_2,\dots,\pi_m)$. 
}

\section{Results}

\subsection{Clustering}
{    
    Prior to text feature extraction using clustering results, we run experiments with the total collections of n-grams found in text corpora. For each type of corpora (human/bot, different languages) 3 million unique n-grams are selected. In order to differentiate bot-generated texts and human-written clusterisations, we study the compactness, separability and noisiness of their clusters.
    \begin{table*}
        \caption{Wilcoxon test p-values for RMSSTD distribution.}
            \label{tab:cluster_wilcoxon}
            \centering
            \begin{tabular}{ccccccc}
            \hline
            {} &  \multicolumn{2}{c}{\textbf{Russian}} &  \multicolumn{2}{c}{\textbf{English}} &  \multicolumn{2}{c}{\textbf{Vietnamese}}\\
            \hline
            {} & {LSTM} & {GPT} & {LSTM} & {GPT} & {LSTM} & {GPT}\\
            \hline
            K-Means & { 5.63e-3 } & { 8.61e-88 } & { 7.47e-4 } & { 4.93e-2 } & { 2.12e-3 } & { 1.50e-2 }\\
            Wishart & { 5.92e-3 } & { 8.15e-28 } & { 4.51e-3 } & { 2.29e-2 } & { 1.32e-5 } & { 9.33e-3	}\\
            \hline
            \end{tabular}
    \end{table*}
    Both the Wishart and K-Means algorithms result in more compact and less separated clusters for bots measured by the RMSSTD and RS metrics \cite{xiong2018clustering}. The three languages share a resemblance --- the clusters for literature corpora are less compact compared to those of bots. The nonparametric Wilcoxon test \cite{wilcoxon1992individual} shows statistically significant differences between RMSSTD distributions of literature and bots corpora: p-values are less than 0.05 (see Table~\ref{tab:cluster_wilcoxon}).
    \begin{figure}[ht]
        \centering
        \includegraphics[width=.5\textwidth]{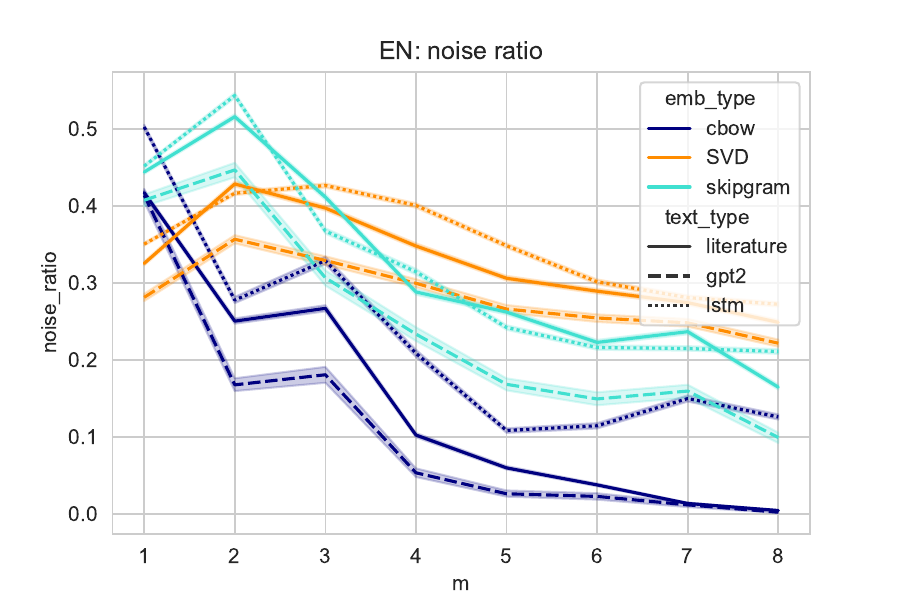}
        \caption{Noise ratio in English data (found with Wishart algorithm on fuzzified data).}
        \label{fig:en_fws_noise}
    \end{figure}
    
    The Wishart clustering algorithm can also be used to find noisy data. We have found that out of all types of texts, those generated by LSTM model are the noisiest, while human written and GPT-generated texts are similar in the noise percentage (see Figure~\ref{fig:en_fws_noise}). We propose a following interpretation for this observation --- the LSTM texts are semantically simpler and the diversity of the texts are mainly achieved by the noise generation.
    \begin{table*}
        \caption{Classification performance (accuracy) with intercluster distance measures.}
            \label{tab:cluster_clf_res}
            \centering
            \begin{tabular}{cccccccc}
            \hline
             &Literature vs. & \multicolumn{2}{c}{\textbf{LSTM+GPT}} & \multicolumn{2}{c}{\textbf{LSTM}} & \multicolumn{2}{c}{\textbf{GPT}}\\
            \hline
            \textbf{Language} & \textbf{Algorithm} & {Train} & {Test} & {Train} & {Test} & {Train} & {Test}\\
            \hline
            English & K-Means & 0.947 & 0.975 & 1.0 & 1.0 & 0.903 & 0.881 \\
            % \hline
            & Wishart & \textbf{0.953} & \textbf{0.975} & 1.0 & 1.0 & 0.904 & 0.881 \\
            % \hline
            & C-Means & 0.943 & 0.970 & 0.999 & 1.0 & 0.897 & 0.921 \\
            % \hline
            & Wishart+Fuzzy & 0.945 & 0.947 & 1.0 & 1.0 & \textbf{0.907} & \textbf{0.94} \\
            \hline
            Russian & K-Means & 0.912 & 0.934 & 0.999 & 1.0 & 0.871 & 0.916 \\
            & Wishart & \textbf{0.937} & \textbf{0.954} & 0.999 & 1.0 & \textbf{0.913} & \textbf{0.944} \\
            & C-Means & 0.882 & 0.894 & 0.999 & 1.0 & 0.838 & 0.857 \\
            & Wishart+Fuzzy & 0.882 & 0.913 & 0.991 & 1.0 & 0.904 & 0.911 \\
            \hline
            Vietnamese & K-Means & 0.862 & 0.903 & 1.0 & 1.0 & 0.887 & 0.881 \\
            % \hline
            & Wishart & 0.902 & 0.896 & 1.0 & 1.0 & 0.893 & 0.900 \\
            % \hline
            & C-Means & 0.887 & 0.893 & 1.0 & 1.0 & 0.871 & 0.871 \\
            % \hline
            & Wishart+Fuzzy & \textbf{0.929} & \textbf{0.942} & 1.0 & 1.0 & \textbf{0.893} & \textbf{0.926} \\   
            \hline
            \end{tabular}
        \end{table*}

    % classification
    Based on these findings, we move on to clustering n-grams for each text in order to extract features. As previous experiments have shown that bots have more compact and less separated clusters, we use inter-cluster distances (average, maximum and minimum) as features. Simple SVC models (separate models for each set of parameters and text types) with L2 regularisation are trained and cross-validated on data subsets (1000 texts for each corpus). Table~\ref{tab:cluster_clf_res} shows the best results for each language. We found that the texts are better distinguished with features extracted from the Wishart algorithm. It is possible that K-Means, as well as its fuzzy variance C-Means, perform worse due to the abstract form of the noisy clusters. It is worth noting that fuzzification improves classification performance on English and Vietnamese texts.
}
\subsection{Entropy-Complexity Plane}
{
    \begin{figure*}[ht]
        \centering
        \includegraphics[width=\linewidth]{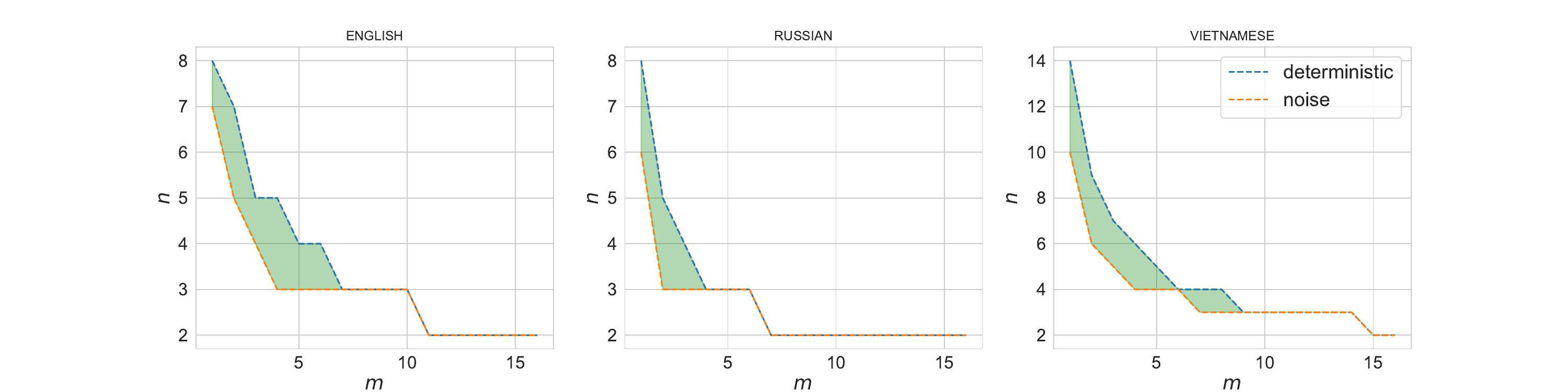}
        \caption{Chaotic area parameter values for English, Russian and Vietnamese data with Skip-gram embeddings.}
        \label{fig:ec_mn}
    \end{figure*}
    
    For certain parameter sets the entropy-complexity measures can fall into noise or deterministic areas, in which it is difficult to identify different types of texts. To account for this nuisance, we first analyse the values of $m$ and $n$ for which the literary texts fall into chaotic area on the entropy-complexity plane (i.e. close to the upper theoretical boundary). Such parameter sets are marked with the green area in Figure~\ref{fig:ec_mn}. Sets below the area border result in texts appearing in noise area, above --- deterministic area. Values differ significantly for each language: longer sequences fall into the chaotic area with values of $n$ varying from 10 to 14 for $m$ = 1 for Vietnamese, whereas for English and Russian the sequences are shorter --- $n$ varies from 7 to 8 and from 6 to 8 accordingly.

    \begin{figure}[ht]
        \centering
        \includegraphics[width=.5\textwidth]{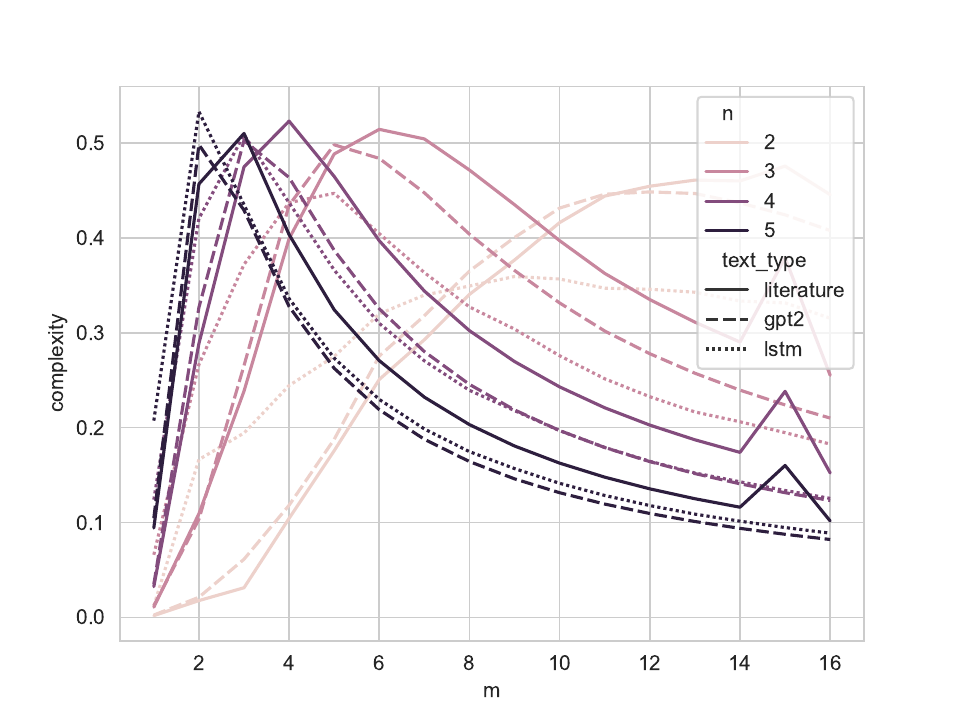}
        \caption{Mean complexity measure on English data.}
        \label{fig:en_c_mn_svd}
    \end{figure}
    On average, the literary texts are more complex, although it is worth noting that for bigrams the more complex texts are LSTM-generated ones (Figure~\ref{fig:en_c_mn_svd}). We believe this happens due to the vast variety of bigrams themselves: more logically coherent texts written by humans or generated by GPT models do not include as many bigrams as the simpler LSTM-generated texts.

    \begin{table*}
            \centering
            \caption{Classification performance (accuracy) based on entropy-complexity measures.}
            \label{tab:ec_clf_res}
            \begin{tabular}{ccccccc}
            \hline
             Literature vs. & \multicolumn{2}{c}{\textbf{LSTM+GPT}} & \multicolumn{2}{c}{\textbf{LSTM}} & \multicolumn{2}{c}{\textbf{GPT}}\\
            \hline
            \textbf{Language} & {Train} & {Test} & {Train} & {Test} & {Train} & {Test}\\
            \hline
            English & 0.937 & 0.965 & 0.999 & 1.0 & 0.997 & 1.0 \\
            Russian & 0.879 & 0.890 & 0.991 & 0.992 & 0.889 & 0.893 \\
            Vietnamese & 0.981 & 0.989 & 1.0 & 1.0 & 0.991 & 0.995 \\
            \hline
            \end{tabular}
    \end{table*}

    For the selected parameter sets we build classification model with entropy and complexity measures as features. Again, for the model we use a simple SVC with L2 regularisation. We originally tried classifying texts with the addition of $m$ and $n$ as numeric features, but such a model only achieved 0.57 accuracy on test set. The models for separate parameter sets perform much better, see Table~\ref{tab:ec_clf_res} for the best models. LSTM texts are well separated on the entropy-complexity plane, a simple SVC achieves 100\% accuracy. GPT texts are also distinguished well --- for English and Vietnamese the accuracy is 99\%, for Russian --- 90\%. The binary classification model for both bots achieves highest accuracy on Skip-gram data, $m=1, n=3$ in English; for Russian --- Skip-gram, $m=1, n=8$; for Vietnamese --- SVD, $m=3, n=3$.

}
\section{Conclusions and further directions}
{
    In order to differentiate generated texts from human literature, we have employed different techniques, such as crisp and fuzzy clustering and entropy-complexity plane construction. We have found that these methods, supplemented by a careful parameter selection, can be used to obtain features with significant differences for different text types. We are therefore able to build robust identification algorithms without prior knowledge of bot-model architecture. The final classification models achieve up to 99\% accuracy for English and Vietnamese data and 94\% for Russian. These methods do not require a lot of labelled data and thus can be easily downstreamed to other tasks, such as fraud detection. As a possible future direction for this work, we also propose an analysis of the methods of this research in application to other languages of varying language families.
}
\section*{Acknowledgements}
{
This research was supported in part through computational resources of HPC facilities at HSE University \cite{kostenetskiy2021hpc}. The authors would also like to thank the HSE AI Center for the support throughout the research process.
}
\bibliographystyle{splncs04}
\bibliography{custom_premi}
\end{document}